\documentclass{article}
\usepackage{amssymb}
\usepackage{amsmath}
\usepackage{tikz}
\usepackage{graphicx}
\usepackage{color}
\usepackage{xcolor}
\usepackage{transparent}
\usepackage{hyperref}
\usepackage{rotating}

\providecommand{\keywords}[1]{\textbf{\textit{Keywords: }} #1}

\DeclareMathOperator*{\argmin}{argmin}

\DeclareMathOperator*{\diag}{diag}

\title{Tensor Methods and Recommender Systems\footnotemark[2]}
\author{Evgeny Frolov\footnotemark[3] \and Ivan Oseledets\footnotemark[3]\ \footnotemark[5]}

\date{}
\DeclareGraphicsExtensions{.pdf, .png}
\graphicspath{ {figures/} }

\begin{document}
\footnotetext[3]{Skolkovo Institute of Science and Technology,
Nobel St.~3, Skolkovo Innovation Center, Moscow, 143025
Moscow Region, Russia; \{i.oseledets,evgeny.frolov\}@skolkovotech.ru}
\footnotetext[5]{Institute of Numerical Mathematics Russian Academy of Sciences,
Gubkina St. 8, 119333 Moscow, Russia}
\footnotetext[2]{This work was supported by Russian Science Foundation grant 14-1100659}
\maketitle

\definecolor{c41719c}{HTML}{FF7F00}

\begin{abstract}
A substantial progress in development of new and efficient tensor factorization techniques has led to an extensive research of their applicability in recommender systems field.
Tensor-based recommender models push the boundaries of traditional collaborative filtering techniques by taking into account a multifaceted nature of real environments, which allows to produce more accurate, situational (e.g. context-aware, criteria-driven) recommendations.
Despite the promising results, tensor-based methods are poorly covered in existing recommender systems surveys. This survey aims to complement previous works and provide a comprehensive overview on the subject.
To the best of our knowledge, this is the first attempt to consolidate studies from various application domains, which helps to get a notion of the current state of the field. We also provide a high level discussion of the future perspectives and directions for further improvement of tensor-based recommendation systems.

%
%
%
%
%
%
%
%
\end{abstract}


\section{Introduction}
\label{sec:Intro}
We live in the era of data explosion and information overload. Managing it would be impossible without the help of intelligent systems that can process and filter huge amounts of information much faster than humans.
The need for such systems was already recognized in late 1970s in the Usenet, a distributed discussion platform, founded at Duke University. One of its goals was to help users to maintain numerous posts by grouping them into newsgroups.
However, an active research on the topic of information filtering started in 1990s. The general term Recommender Systems (RS) was brought to the academia in the mid-90's with works of Resnick, Hill, Shardanand and Maes \cite{Adomavicius2005}
and was preceded by several famous projects:
Tapestry, Lotus Notes, GroupLens \cite{brusilovsky2008social}.
A significant boost in RS research started after a famous Netflix prize competition with \$1 million award for the winners, announced back in 2006. This has not only attracted a lot of attention from scientists and engineers, but also depicted the great interest from an industry.

Conventional RS deal with two major types of entities which are typically users (e.g. customers, consumers) and items (e.g. products, resources). Users interact with items by viewing or purchasing them, assigning ratings, leaving text reviews, placing likes or dislikes, etc. These interactions, also called events or transactions, create an observation history, typically collected in a form of transaction/event log, that reflects the relations between users and items. Recognizing and \emph{learning} these relations in order to predict new possible interactions is one of the key goals of RS.

As we will see further, the definition of entities is not limited to users and items only. Entities can be practically of any type as long as predicting new interactions between them may bring a valuable knowledge and/or help to make better decisions. In some cases entities can be even of the same type, like in the task of predicting new connections between people in a social network or recommending relevant paper citations for a scientific papers.

Modern recommender models may also have to deal with more than 2 types of entities within a single system. For instance, users may want to assign tags (e.g. keywords) to the items they like.
Tags become the third type of entity, that relates to both users and items, as it represents the user motivation and clarifies items relevance (more on that in Section \ref{subsec:tagrec}).
Time can be another example of an additional entity, as both user preferences and items relevance may depend on time (see Section \ref{subsec:temporal}). Taking into account these multiple relations between several entities typically helps to provide more relevant, dynamic and situational recommendations. It also increases complexity of RS models, which in turn brings new challenges and opens the door for new types of algorithms, such as tensor factorization (TF) methods.

The topic of building a production-ready recommender system is very broad and includes not only algorithms but also concerns a lot about business logic, dataflow design, integration with infrastructure, service delivery and user experience. This also may require a specific domain knowledge and always needs a comprehensive evaluation. Speaking about the latter, the most appropriate way of assessing RS quality is an online A/B testing and massive user studies \cite{Herlocker2004, ekstrand2011collaborative, knijnenburg2015evaluating}, which are typically not available right at hand in academia. In this work we will only touch mathematical and algorithmic aspects which will be accompanied with examples from various application domains.

The rest of the survey is divided into the following parts: Sections \ref{sec:rs_intro} and \ref{sec:rs_chal} cover general concepts and major challenges in RS field; Section \ref{sec:tensor_intro} gives a brief introduction to tensor-related concepts, that are essential for understanding how tensors can be used in RS models; Section \ref{sec:tensor_models} contains a comprehensive overview of various tensor-based techniques with examples from different domains; Section \ref{sec:conclusion} concludes the review and provides thoughts on possible future directions.

\section{Recommender systems at a glance}
\label{sec:rs_intro}
Let us consider without loss of generality the task of product recommendations. The main goal of this task is, given some prior information about users and items (products), try to predict what particular items will be the most relevant to a selected user.
The relevance is measured with some relevance score (or utility) function $f_R$ that is estimated from the user feedbacks. More formally,
\begin{equation}
\label{eq:utility}
    f_R: User \times Item \rightarrow Relevance\,Score,
\end{equation}
where $User$ is a domain of all users, $Item$ is a domain of all items.
The feedback can be either explicit or implicit, depending on whether it is directly provided by a user (e.g. ratings, likes/dislikes, etc.) or implicitly collected through an observation of his/her actions (e.g. page clicks, product purchases, etc.).

The type of prior information available in RS model defines what class of techniques will be used for building recommendations. If only an observation history of interactions can be accessed, then this is a task for collaborative filtering (CF) approach. If RS model uses intrinsic properties of items as well as profile attributes of users in order to find the best matching (\emph{user, item}) pairs, then this is a content-based (CB) approach.


The complete overview of RS methods and challenges is out of scope of this survey and for a deeper introduction
we refer the reader to \cite{ekstrand2011collaborative, Bobadilla2013, Shi2014, Ricci2015}.

\subsection{Content-based filtering}
As already mentioned, the general idea behind the CB approach is to use some prior knowledge about users preferences and items' properties in order to generate the most relevant recommendations.
One of its main advantages is the ability to alleviate the cold start problem (see Section \ref{subsec:cold_start}) as long as all the needed content information is collected. Recommendations can be produced instantly even for those items that were never recommended to any user before.

This approach also has a number of issues, among which are the limited content analysis, over-specialization and high sensitivity to users input \cite{Adomavicius2005, Lops2011}.
The key drawback from practical viewpoint is the difficulty of gathering descriptive and thorough properties of both items and users. This can be either manually done by humans, i.e. with help of users and/or domain experts, or extracted automatically with data processing tools.
The former method is usually very time consuming and requires considerable amount of work before RS can be built up. The latter method is highly dependent on information retrieval (IR) algorithms and is not always accurate or even possible.

\subsection{Collaborative filtering}
In contrast to CB filtering, CF does not require any specific knowledge about users or items and only uses prior observations of users' collective behavior in order to build new recommendations. The class of CF techniques is generally divided into two categories: memory-based and model-based methods \cite{Bobadilla2013, Adomavicius2005a}.

\subsubsection{Memory-based collaborative filtering}
\label{subsec:knn}
A widely used and very popular approach in this category is based on \emph{k Nearest Neighbours} (kNN) algorithm \cite{hastie2005elements}. It finds relevance scores for any (\emph{user, item}) pair by calculating contributions from its neighbors. The neighborhood is typically determined by a similarity between either users (user-based approach) or items (item-based approach) \cite{Schafer2007} in terms of some similarity measure. This is also called a similarity-based approach.
In its simplest implementation, the method requires to store in memory all prior information about user-item interactions in order to make predictions.

Performance of the similarity models may be greatly impacted by a selected measure of similarity (or a distance measure). Cosine similarity, Jaccard index, Pearson correlation, Okapi BM25 \cite{parra2009collaborative}
are a few examples of possible choices.
Even though the pure similarity-based models may give a good recommendations quality in some application domains, factorization models (see Section \ref{subsec:model_based}) are better suited for large-scale problems often met in practice, delivering high performance and high quality recommendations \cite{Konstan2012, Bobadilla2013}.
\subsubsection{Model-based collaborative filtering}
\label{subsec:model_based}
In the model-based approach a predictive model is generated from a long enough history of observations and uses collective behavior of the crowd (a ``wisdom of crowds'') in order to extract general behavioral patterns. One of the most successful model-based approaches is a matrix factorization (MF). The power of factorization models comes from the ability to embed users and items as vectors in a lower dimensional space of latent (or hidden) features (see Section \ref{subsec:reduce}).
These models represent both users' preferences and corresponding items' features in a unified way so that the relevance score of the user-item interaction can be simply measured as an inner product of their vectors in the latent feature space.


As it follows from the description, both CF and CB tackle the problem of building relevant recommendations in very different ways and have their own sets of advantages and disadvantages. Many successful RS use \emph{hybrid} approaches, that combine the advantages of both methods within a single model \cite{Koren2008, Burke2007}.

\section{Challenges for recommender systems}
\label{sec:rs_chal}
Building high quality RS is a complex problem, that involves not only a certain level of scientific knowledge but also greatly relies on an experience, passed from an industry and facing the real world implementations.
This topic is also very broad and we will briefly discuss only the most common challenges, that are closely related to an initial model design and its algorithmic implementations.

\subsection{Cold-start}
\label{subsec:cold_start}
Cold-start is the problem of handling new entities, that concerns with both users and items \cite{ekstrand2011collaborative}. When a new user is introduced to the system we usually know little or nothing about the user preferences and thus it makes it difficult or impossible to predict any interesting items for him or her. Similar problem arises when a new item appears in a product catalog. If an item has no content description or it was not rated by any user it will be impossible to build recommendations with this item.


\subsection{Missing values}
\label{subsec:missing}
Users typically engage with only a small subset of items and considerable amount of possible interactions stays unobserved. Excluding the trivial case of the lack of interest in specific items, there may be some other reasons for not interacting with them. For example, users may be simply unaware of existing alternatives for the items of their choice. Finding out those reasons helps to make better predictions and, of course, is a part of RS task. However, high level of uncertainty may bring an undesirable bias against unobserved data or even prevent RS models from learning representative patterns, resulting in low recommendations quality.

There are several commonly used techniques, that help to alleviate these issues and improve RS quality. In MF case, simple regularization may prevent the undesired biases. Another effective technique is to assign some non-zero weights to the missing data, instead of completely ignoring it \cite{Hu2008}. In hybrid models a content information can be used in order to pre-process observations and assign non-zero relevance scores to some of the unobserved interactions (sometimes called as \emph{sparsity smoothing}). This new data is then fed into standard CF procedure. Data clustering is another effective approach that is typically used to split the problem into a number of subproblems of smaller size with more connected information.
Nevertheless, in case of a particular MF method, based on Singular Value Decomposition (SVD) \cite{golub1970singular}, simply imputing zero relevance scores for an unobserved values may produce better results \cite{Cremonesi2010, Lee2016a}. Additional smoothing can be achieved in that case with help of a \emph{kernel trick} \cite{Shawe-Taylor2004}. Other missing value imputation techniques based on various data averaging and normalization methods are also possible \cite{ekstrand2011collaborative}.
As we will see in Section \ref{sec:tensor_models}, all of these techniques are valid in TF case as well.


\subsection{Implicit feedback}
\label{subsec:intro_implicit}
In many real systems users are not motivated or not technically equipped to provide any information about their actual experience after interacting with an item. Hence, user preferences can only be inferred from an implicit feedback, which may not necessarily reflect the actual user taste or even tell with guarantees whether the user likes an item or dislikes it \cite{Hu2008}.


\subsection{Model evaluation}
\label{subsec:eval}
Without a well designed evaluation workflow and an adequate quality measures it is impossible to build a reliable RS model that behaves equally well in both laboratory and production environments.
Moreover, there are many aspects of a model assessment beyond recommendations accuracy, that are related to both user experience and business goals. This can include metrics like \emph{coverage, diversity, novelty, serendipity} (see \cite{Shani2011} for explanations) and indicators such as total generated revenue or average revenue per user session.
This is still an open and ongoing research problem as it is not totally clear what are the most relevant and informative offline metrics and how to align them with the real online performance.

As mentioned in the beginning of Section \ref{sec:rs_chal}, the most reliable evaluation of RS performance is an online testing and user studies. Researchers typically do not have an access to a production systems so a number of offline metrics (mostly borrowed from IR field), became very popular. The most important among them are the relevance metrics: precision, recall, F1-score
and the ranking metrics: normalized discounted cumulative gain (NDCG), mean average precision (MAP), mean reciprocal rank (MRR), area under the ROC curve (AUC). These metrics may to some extent simulate a real environment, and in same cases have strong correlation with business metrics (e.g. recall and clickthrough rates (CTR) \cite{Hidasi2015}).

It is also important to emphasize that while there are some real-world systems that target a direct prediction of a relevance score (e.g. rating), in most cases the main goal of RS is to build a good ranked list of items (top-$n$ recommendations task).
This imposes some constraints on the evaluation techniques and model construction. It might be tempting to use and optimize for error-based metrics like root mean squared error (RMSE) or mean absolute error (MAE) due to their simplicity. However, good performance in terms of RMSE does not guarantee a good performance on generating a ranked list of top-$n$ recommendations \cite{ekstrand2011collaborative}.
In other words, the predicted relevance score may not align well with the perceived quality of recommendations.

\subsection{Reproducible results}
The problem of reproducibility is closely related to recommendations quality evaluation. Careful design of evaluation procedures is critical for fair comparison of various methods.
However, independent studies show that in controlled environments it is problematic to get consistent evaluation results even for the same algorithms on fixed datasets but within different platforms \cite{Said2014}.

Situation gets even worse, taking into account that many models, that tackle similar problems, use different datasets (sometimes not publicly available), different data pre-processing techniques \cite{Doerfel2016} or different evaluation metrics.
In order to avoid unintended biases, we will focus mostly on the description of the key features of existing methods rather than on a side-by-side comparison of quantitative results.


\subsection{Real-time recommendations}
\label{subsec:online}
A high quality RS are expected not only to produce relevant recommendations but also respond instantly to the system updates, such as new (or unrecognized) users, new items or new feedbacks \cite{Konstan2012}. Satisfying the latter requirement highly depends on the implementation: the predictive algorithms must have low computational complexity for producing new recommendations and take into account the dynamic nature of a real environments.
Recomputation of the full RS model in order to include the new entities may take prohibitively long time and the user may never see a recommendation before he or she leaves. This means that RS application should be capable of making incremental updates and also be able to provide instant recommendations at a low computational cost outside of the full model recomputation loop. A number of techniques has been developed to fulfill these requirements for the MF case \cite{furnas1988ir, Zha1999, brand2002incremental}. As it will be shown in Section \ref{subsec:tagrec}, these ideas can be also applied in the TF case.

\subsection{Incorporating context information}
\label{subsec:incontext}
\begin{figure}[bt]
\noindent\centering{
\includegraphics[width=110mm]{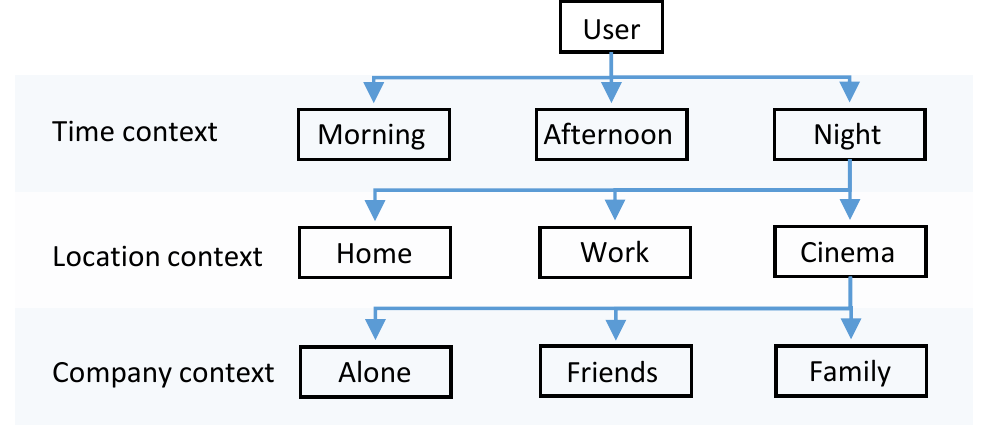}
}
\caption{Examples of contextual information.}
\label{fig:context}
\end{figure}

In the real world scenarios interactions between users and items exhibit a multifaceted nature. User preferences are typically not fixed and may change with respect to a specific situation.
For example, buyers may prefer different goods depending on the season of the year or time of the day. A user may prefer to watch different movies when alone or with a company of friends.
We will informally call these situational aspects, that shape user behavior, a contextual information or a context for short (see Figure \ref{fig:context}). Another examples of context are location, day of week, mood, the type of a user's electronic device, etc. Essentially, it can be almost anything \cite{Bazire2005, Dourish2004}.

Context-aware recommender systems (CARS) can be built with 3 distinct techniques
\cite{Adomavicius2011}: contextual prefiltering, where a separate model is learned for every context type; contextual postfiltering, where adjustments are performed after a general context-unaware model was built; and
contextual modelling, where context becomes an essential part of the training process.
The first two techniques may lose information about the interrelations within a context itself.
Contextual modelling, in turn, extends the dimensionality of the problem and promotes multirelational aspect into it. Therefore it is likely to provide more accurate results \cite{Karatzoglou2010}. Following \eqref{eq:utility}, we can formalize it as follows:
\begin{equation}
\label{eq:NDutility}
    f_R: User \times Item \times Context_1 \times \ldots \times Context_N \rightarrow Relevance\,Score,
\end{equation}
where $Context_i$ denotes one of $N$ contextual domains and the overall dimensionality of the model is $N\!+\!2$.

As we will see further, TF models fit perfectly into the concept of CARS. With a very broad definition of context, tensor-based methods turn into a flexible tool, that allows to naturally model very interesting and non-trivial setups, where the concept of context goes beyond a typical connotation.

As a precaution, it should be noted that a nonspecifity of a context may lead to an interpretability problems. Using a general definition of a context, a content information such as user profile attributes (e.g. age, gender) or items properties (e.g. movie genre or product category) can also be regarded as some type of context(see, for example, \cite{Karatzoglou2010}, where age and gender are used to build new context dimensions).
However, in practice, especially for TF models, this mixing is typically avoided \cite{Woerndl2007, Rettinger2012}. One of the possible reasons is a deterministic nature of content information in contrast to what is usually denoted as a context. Similarly to MF techniques, TF reveals new unseen associations (see Section \ref{subsec:reduce}) which in the case of deterministic attributes may be hard to interpret. It is easy to see in the following example.

For a triplet (\emph{user, movie, gender}) the movie rating may be associated with only one of two possible pairs of (\emph{user, gender}), depending on the actual user's gender. However, once a reconstruction (with help of some TF technique) is made, a non-zero value of rating may now pop-up for both values of gender. The interpretation of such an association may become tricky and highly depends on initial problem formulation.


\section{Introduction to tensors}
\label{sec:tensor_intro}

In this section we will only briefly introduce some general concepts needed for better understanding of further material. For a deeper introduction to the key mathematical aspects of multilinear algebra and tensor factorizations we refer the reader to \cite{Kolda2009, Comon2014, Grasedyck2013}.
As in the case of MF in RS, TF produces a predictive model by revealing patterns from the data. The major advantage of a tensor-based approach is the ability to take into account a multifaceted nature of user-item interactions.

\subsection{Definitions and notations}
We will regard an array of numbers with more than 2 dimensions as a \emph{tensor}. This is a natural extension of matrices to a higher order case. A tensor with $m$ distinct dimensions or \emph{modes} is called an $m$-way tensor or a tensor of order $m$.

 Without loss of generality and for the sake of simplicity we will start our considerations with a 3rd order tensors to illustrate some important concepts. We will denote tensors with calligraphic capital letters, e.g. $\mathcal{T} \in \mathbb{R}^{M \times N \times K}$ stands for a 3rd order tensor of real numbers with dimensions of sizes $M, N, K$. We will also use a compact form $\mathcal{T} = [t_{ijk}]_{i, j, k=1}^{M, N, K}$, where $t_{ijk}$ is an element or entry at position $(i, j, k)$, and will assume everywhere in the text the values of the tensor to be real.

\paragraph{Tensor fibers.} A generalization of matrix rows and columns to a higher order case is called a \emph{fiber}. Fiber represents a sequence of elements along a fixed mode when all but one indices are fixed. Thus, a mode-1 fiber of a tensor is equivalent to a matrix column, a mode-2 fiber of a tensor corresponds to a matrix row. A mode-3 fiber in a tensor is also called a tube.

\paragraph{Tensor slices.} Another important concept is a tensor \emph{slice}. Slices can be obtained by fixing all but two indices in a tensor, thus forming a two-dimensional array, i.e. matrix. In a third order tensor there could be 3 types of slices: horizontal, lateral, and frontal, which are denoted as $\mathcal{T}_{i::}, \mathcal{T}_{:j:}, \mathcal{T}_{::k}$ respectively.

\begin{figure}[bt]
\noindent\centering{
\includegraphics[width=110mm]{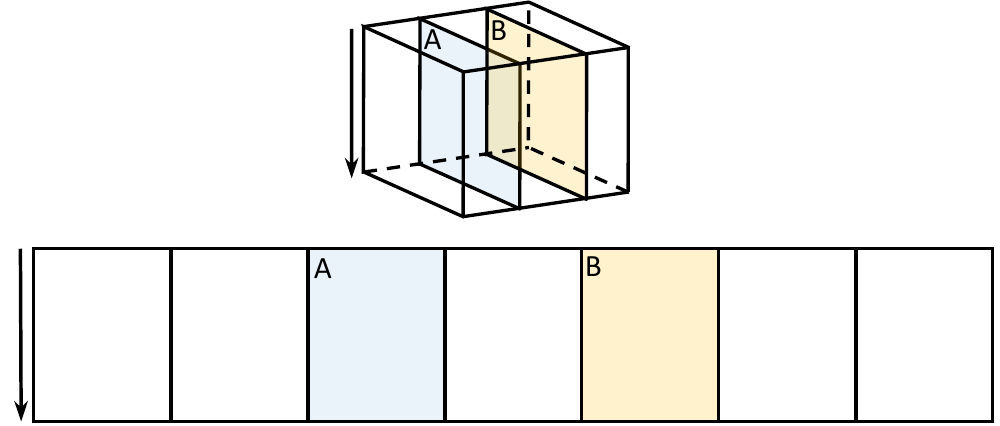}
}
\caption{Tensor of order 3 (top) and its unfolding (bottom). Arrow denotes the mode of matricization.}
\label{fig:matricize}
\end{figure}

\paragraph{Matricization.}
Matricization is a key term in tensor factorization techniques. This is a procedure of reshaping a tensor into a matrix. Sometimes it is also called unfolding or flattening. We will follow the definition introduced in \cite{Kolda2009}.
The $n$-mode matricization of a tensor $\mathcal{T} \in \mathbb{R}^{M \times N \times K}$ arranges the mode-$n$ fibers to be the columns of the resulting matrix (see Figure \ref{fig:matricize}). For the 1-mode matricization $T_{(1)}$ the resulting matrix size is $M \times (N K)$, for the 2-mode matricization $T_{(2)}$ the size is $N \times (M K)$ and the 3-mode matricization $T_{(3)}$ has the size $K \times (M N)$.
In the general case of an $m$-th order tensor $\mathcal{T} \in \mathbb{R}^{I_1\times I_2 \times \dots \times I_m}$  the $n$-mode matricization $T_{(n)}$ will have the size $I_n \times (I_1 I_2 \dots I_{n-1} I_{n+1} \dots I_m)$.
For the corresponding index mapping rules we refer the reader to \cite{Kolda2009}.


\paragraph{Diagonal tensors.}
Another helpful concept is a diagonal tensor. A tensor $\mathcal{T} \in \mathbb{R}^{I_1 \times I_2 \times \dots \times I_m}$ is called diagonal if $t_{i_1 i_2 \dots i_m} \neq 0$ only if $i_1\!=\!i_2\!=\!\ldots\!=~\!i_m$.
This concept helps to build a connection between different kinds of tensor  decompositions.


\subsection{Tensor Factorization techniques}
The concept of TF can be better understood via an analogy with MF. For this reason we will first introduce a convenient notation and representation for the MF case and then generalize it to a higher order.

\subsection{Dimensionality reduction as a learning task}
\label{subsec:reduce}
Let us first start with SVD, as it helps to illustrate some important concepts and also serves as a workhorse for certain TF techniques.
Any matrix $A \in \mathbb{R}^{M \times N}$ can be represented in the form:
\begin{equation}
\label{eq:svd}
A = U \Sigma V^{T},
\end{equation}
where $U \in \mathbb{R}^{M \times K}$ and $V \in \mathbb{R}^{N \times K}$ are orthogonal matrices, $\Sigma = \diag(\sigma_1, \dots, \sigma_K)$ is a diagonal matrix of non-negative singular values $\sigma_1 \geq \ldots \geq \sigma_K$ and $K = \min(M, N)$
is a \emph{rank} of SVD. According to the Eckart-Young theorem \cite{eckart_young}, the truncated SVD of rank $r < K$ with $\sigma_{r+1}, \dots, \sigma_{K}$ set to 0 gives the best rank-$r$ (also called low-rank) approximation of matrix $A$. This has a number of important implications for RS models.

A typical user-item matrix in RS represents a snapshot of real ``noisy'' data and it is practically never of low-rank.
However, the collective behavior has some common patterns which yield a low-rank structure and the real data can be modelled as:
\begin{displaymath}
    R = A_r + E,
\end{displaymath}
where $E$ is a ``noise'' and $A_r$ is a rank-$r$ approximation of data matrix $R$. The task of building a recommendation model translates into the task of recovering $A_r$ (or equivalently, minimizing the noise $E$). For illustration purposes here we assume that missing data in $R$ is replaced with zeroes (other ways of dealing with missing values problem are briefly described in Section \ref{subsec:missing}). Despite the simplicity of the assumption this approach is known to serve as a strong baseline \cite{Cremonesi2010, Lee2016a}. The Eckart-Young theorem states, that an optimal solution to the resulting optimization task
\begin{equation}
\label{eq:optimat}
    \min_{A_r} \|R - A_r\|^2
\end{equation}
is given by the truncated SVD:
\begin{equation}
\label{eq:lowrank}
    R \approx A_r = U_r \Sigma_r V_r^{T}.
\end{equation}
Here and further in the text we use $\| \cdot \|$ to denote Frobenius norm (both for matrix and tensor case), if not specified otherwise.

In terms of RS, factor matrices $U_r$ and $V_r$, learned from observations,  represent an embedding of users and items into the reduced latent space with $r$ latent features.
The dimensionality reduction produces a ``denoised picture'' of data, it reveals a hidden (latent) structure that describes the relations between users and items.
With this latent representation some previously unobserved interactions can be uncovered and used to generate recommendations. This idea can be extended to a case of higher order relations between more than 2 entities and that is where a tensor factorizations techniques come into play.

From now on we will omit the subscript $r$ in the equations for both matrix and tensor factorizations, e.g. we will denote a factor matrix $U_r$ simply as $U$ and likewise for other factor matrices. Let us also rewrite \eqref{eq:svd} in the form, that is useful for further generalization to higher order case:
\begin{equation}
\label{eq:n-mode}
A = \Sigma \times_1 U \times_2 V,
\end{equation}
where $\times_n$ is an \emph{n-mode product} which is typically defined for product between high order tensor and matrix.
In matrix case the $n$-mode product between two matrices $A$ and $B$ has the following form (assuming they are conformable):
\begin{displaymath}
(A \times_1 B)_{ij} = \sum_{k} a_{ki} b_{jk}, \quad
(A \times_2 B)_{ij} = \sum_{k} a_{ik} b_{jk}.
\end{displaymath}
In a more general case each resulting element of an $n$-mode product between tensor $\mathcal{T}$ and matrix $U$ is calculated as follows \cite{Kolda2009}:
\begin{equation}
\label{eq:n-mode-general}
    (\mathcal{T} \times_n U)_{i_1 \dots i_{n-1} j i_{n+1} \dots i_m} = \sum_{i_n} t_{i_1 i_2 \dots i_m} u_{ji_n}.
\end{equation}
For the same purpose of further generalization we will rewrite \eqref{eq:n-mode} in 2 other forms, the index form:
\begin{displaymath}
    a_{ij} = \sum_{\alpha=1}^r \sigma_\alpha u_{i\alpha} v_{j\alpha},
\end{displaymath}
and a sum of rank-1 terms:
\begin{equation}
    \label{eq:rank1}
    A = \sum_{\alpha=1}^r \sigma_\alpha \boldsymbol{u}_\alpha \otimes \boldsymbol{v}_\alpha,
\end{equation}
where $\boldsymbol{u}_\alpha, \boldsymbol{v}_\alpha$ denote columns of the factor matrices, e.g. $U = [\boldsymbol{u}_1 \dots \boldsymbol{u}_r], V = [\boldsymbol{v}_1 \dots \boldsymbol{v}_r]$ and $\otimes$ denotes the vector outer product (or dyadic product).

In the tensor case we will also be interested in the task of learning a factor model from a real observations data $\mathcal{Y}$. This turns into a dimensionality reduction problem that gives a suitable (not necessarily the best in terms of error-based metrics) approximation:
\begin{displaymath}
    \mathcal{Y} \approx \mathcal{T},
\end{displaymath}
where $\mathcal{T}$ is calculated with help of some of the tensor decomposition methods, described further. We will keep this notation throughout the text, e.g. $\mathcal{Y}$ will always be used to denote a real data and $\mathcal{T}$ will always be used to represent the reconstructed model, learned from $\mathcal{Y}$.

\subsubsection{Candecomp/Parafac}
\label{subsec:cp}
The most straightforward way of extending SVD to higher orders is to add new factors in \eqref{eq:rank1}. In the third order case this will have the following form:
\begin{equation}
    \label{eq:cp}
    \mathcal{T} = \sum_{\alpha=1}^r \lambda_\alpha \boldsymbol{u}_\alpha \otimes \boldsymbol{v}_\alpha \otimes \boldsymbol{w}_\alpha,
\end{equation}
where each summation component $\boldsymbol{u}_\alpha \otimes \boldsymbol{v}_\alpha \otimes \boldsymbol{w}_\alpha$ is a \emph{rank-1} tensor. We can also equivalently rewrite \eqref{eq:cp} in a more concise notation:
\begin{equation}
\label{eq:cpbr}
\mathcal{T} = [\![ \pmb{\lambda}; U, V, W ]\!],
\end{equation}
where $\pmb{\lambda}$ is a vector of length $r$ with elements $\lambda_1 \geq \ldots \geq \lambda_r > 0$ and $U \in \mathbb{R}^{M \times r}$, $V \in \mathbb{R}^{N \times r}$, $W \in \mathbb{R}^{K \times r}$ defined similarly to \eqref{eq:rank1}.
The expression assumes that factors $U, V, W$ are normalized. As we will see further, in some cases values of $\pmb{\lambda}$ can have a meaningful interpretation. However, in general, the assumption can be safely omitted, which yields:
\begin{equation}
    \label{eq:cp1}
    \mathcal{T} = [\![U, V, W ]\!] \equiv \sum_{\alpha=1}^r \boldsymbol{u}_\alpha \otimes \boldsymbol{v}_\alpha \otimes \boldsymbol{w}_\alpha,
\end{equation}
or in the index from:
\begin{equation}
    \label{eq:cpidx}
    t_{ijk} = \sum_{\alpha=1}^{r} \, u_{i\alpha} \, v_{j\alpha} \, w_{k\alpha}.
\end{equation}
The right-hand side of \eqref{eq:cp1} gives an approximation of real observations data and is called Candecomp/Parafac (CP) decomposition of a tensor $\mathcal{Y}$. Despite being similar to \eqref{eq:rank1} formulation, there is a number of substantial differences in the concepts of tensor rank and low-rank approximation, thoroughly explained in \cite{Kolda2009}. Apart from technical considerations, an important conceptual difference is that there is no higher order extension of the Eckart-Young theorem (mentioned in the beginning of Section \ref{subsec:reduce}), i.e. if an exact low-rank decomposition of $\mathcal{Y}$ with rank $r'$ is known, then its truncation to the first $r < r'$ terms may not give the best rank-$r$ approximation. Moreover, the optimization task in terms of low-rank approximation is ill-posed \cite{Silva} which is likely to lead to numerical instabilities and issues with convergence, unless additional constraints on factor matrices (e.g. orthogonality,  non-negativity, etc.) are imposed.


\subsubsection{Tucker decomposition}
\label{subsec:tucker}
A stable way of extending SVD to a higher order case is to transform the diagonal matrix $\Sigma$ from \eqref{eq:n-mode} into a third order tensor $\mathcal{G}$ and add an additional mode-3 tensor product (defined by \eqref{eq:n-mode-general}) with a new factor matrix $W$:
\begin{equation}
    \label{eq:tucker}
    \mathcal{T} = [\![\mathcal{G}; U, V, W]\!] \equiv \mathcal{G} \times_1U\times_2V\times_3W,
\end{equation}
where $U \in \mathbb{R}^{M \times r_1}, V \in \mathbb{R}^{N \times r_2}, W \in \mathbb{R}^{K \times r_3}$ are orthogonal matrices, having similar meaning of the latent feature matrices as in the case of SVD. Tensor $\mathcal{G} \in \mathbb{R}^{r_1 \times r_2 \times r_3}$ is called a core tensor of the TD and a tuple of numbers ($r_1, r_2, r_3$) is called a multilinear rank. The decomposition is not unique, however the optimization problem with respect to multilinear rank is well-posed.
Note, that if tensor $\mathcal{G}$ is diagonal with all ones on its diagonal, than the decomposition turns into CP. In the index notation TD takes the following form:
\begin{equation}
\label{eq:tuckidx}
t_{ijk} = \sum_{\alpha,\beta,\gamma=1}^{r_1, r_2, r_3}{g_{\alpha\beta\gamma}} \, u_{i\alpha} \, v_{j\beta} \, w_{k\gamma}.
\end{equation}

The definition of TD is not restricted to have 3 modes only. Generally, the number of modes is not limited, however storage requirements depend exponentially on the number of dimensions (see Table \ref{tab:storage}), which is often referred as a \emph{curse of dimensionality}. This imposes strict limitations on the number of modes for many practical cases, whenever more than 4 entities are modelled in a multilinear way (e.g. \emph{user}, \emph{item}, \emph{time}, \emph{location}, \emph{company} or any other context variables, see Figure 1). In order to break the curse of dimensionality, a number of efficient methods has been developed recently, namely Tensor Train (TT) \cite{Oseledets2011} and Hierarchical Tucker (HT) \cite{grasedyck2010hierarchical}. However, we are not aware of any published results related to TT- or HT-based implementations in RS.

\begin{table}
\centering{
\begin{tabular}{l|cccc}
{} &     CP &           TD &       TT &   HT \\
\hline
storage &  $dnr$ &  $dnr + r^d$ &  $dnr^2$ &  $dnr + dr^3$ \\
\end{tabular}
}
\caption{Storage requirements for different TF methods.
For the sake of simplicity, this assumes a tensor with $d$ dimensions of equal size $n$
and all ranks (or rank in case of CP) of a tensor decomposition set to $r$.}
\label{tab:storage}
\end{table}



\subsubsection{Optimization algorithms}
Let us start from the simplest form of an optimization task where the objective $J$ is defined by a loss function $L$ as follows:
\begin{equation}
\label{eq:simpleopt}
    J(\theta) = L(\mathcal{Y}, \mathcal{T}(\theta)),
\end{equation}
where $\theta$ denotes model parameters, i.e.  $\theta := (U, V, W)$ for CP-based models and $\theta := (\mathcal{G}, U, V, W)$ in case of TD.
The optimization criteria takes the following form:
\begin{equation}
    \label{eq:argminj}
    \theta^* = \argmin_{\theta} J(\theta),
\end{equation}
where $\theta^*$ defines an optimal set of model parameters which are to be used to generate recommendations.

It is convenient to distinguish the three general categories of optimization objectives that lead to different ranking mechanisms:  \emph{pointwise}, \emph{pairwise} and \emph{listwise} \cite{chapelle2010}.
Pointwise objective depends on a pointwise loss function
between the observations $y_{ijk}$ and the predicted values $t_{ijk}$. In case of a square loss the total loss function has a similar to \eqref{eq:optimat} form:
\begin{equation}
\label{eq:pointwise}
L(\mathcal{Y}, \mathcal{T}) = \|\mathcal{Y} - \mathcal{T}\|^2.
\end{equation}

Pairwise objective depends on a pairwise comparison of the predicted values and penalizes those cases when their ordering does not correspond to the ordering of observations. The total loss can take the following form in that case:
\begin{displaymath}
L(\mathcal{Y}, \mathcal{T}) = \sum_{i, j} \sum_{k, k': y_{ijk} > y_{ijk'}} l(t_{ijk} - t_{ijk'}),
\end{displaymath}
where $l(x)$ is a pairwise loss function that decreases with the increase of $x$.

Listwise objective operates over whole sets (or lists) of predictions and observations. The listwise loss function, that can be schematically expressed as $l(\{t_{ijk}\}, \{y_{ijk}\})$, penalizes the difference between the predicted ranking of a given list of items and the ground truth ranking, known from observations.


\paragraph{Pointwise algorithms for TD.}
In case of TD-based model the solution to \eqref{eq:argminj} given \eqref{eq:pointwise} can be found with help of two well-known methods proposed in \cite{DeLathauwer2000}: Higher-Order SVD (HOSVD) \cite{Sun2005, Symeonidis2010, Rafailidis2013} or Higher-Order Orthogonal Iteration (HOOI) \cite{Zhang2014, Symeonidis2015}.

The HOSVD method can be described as a consecutive application of SVD to all 3 matricizations of $\mathcal{Y}$, i.e. $Y_{(1)}, Y_{(2)}, Y_{(3)}$ (assuming that missing data is imputed with zeros). Generally it produces a suboptimal solution to an optimization problem induced by \eqref{eq:simpleopt}, however, it is worse than the best possible solution only by a factor of $\sqrt{d}$, where $d$ is the number of dimensions \cite{hackbusch2012}. Due to its simplicity this method is often used in recommender systems literature.

The HOOI method uses an iterative procedure based on an alternating least squares (ALS) technique, which successively optimizes \eqref{eq:simpleopt}. In practice it may require a small amount of iterations to converge to an optimal solution, but in general it is not guaranteed to find a global optimum \cite{Kolda2009}.
The choice of any of these two methods for particular problem may require additional investigation in terms of both computational efficiency and recommendations quality before the final decision is made.


The orthogonality constraints imposed by TD may in some cases have no specific interpretation. Relaxing these constraints leads to a different optimization scheme, typically based on gradient methods, such as stochastic gradient descent (SGD) \cite{Karatzoglou2010}. The objective in that case is expanded with a regularization term $\Omega(\theta)$:
\begin{equation}
    \label{eq:opt_general}
    J(\theta) = L(\mathcal{Y}, \mathcal{T}(\theta)) + \Omega(\theta),
\end{equation}
which is commonly expressed as follows:
\begin{equation}
    \label{eq:opt_reg}
    \Omega(\theta) = \lambda_G\|\mathcal{G}\|^2 + \lambda_U\|U\|^2 + \lambda_V\|V\|^2 + \lambda_W\|W\|^2, \\
\end{equation}
where $\lambda_G$, $\lambda_U, \lambda_V, \lambda_W$ are regularization parameters and usually \mbox{$\lambda_U = \lambda_V = \lambda_W$}.


\paragraph{Pointwise algorithms for CP.}
As has been noted in Section \eqref{subsec:cp}, CP is generally ill-posed and if no specific domain knowledge could be employed to impose additional constraints, a common approach to alleviate the problem is to introduce regularization similarly to \eqref{eq:opt_reg}:
\begin{equation}
    \label{eq:opt_reg_cp}
    \Omega(\theta) = \lambda_U\|U\|^2 + \lambda_V\|V\|^2 + \lambda_W\|W\|^2, \\
\end{equation}
Indeed, depending on the problem formulation it may also have more complex form both for CP (e.g. as in Section \ref{subsec:BPTF}) and TD models.
In general, regularization allows to ensure convergence and avoid degeneracy (e.g. when rank-1 terms become close to each other by absolute value but their magnitudes go to infinity and have opposite signs \cite{Kolda2009}), however it may lead to a sluggish rate of convergence \cite{navasca2008swamp}.
In practice, however, many problems can still be solved with CP using variations of both ALS \cite{Hidasi2015, Kolda2006} and gradient-based methods.


\paragraph{Pairwise and listwise algorithms.}
Pairwise and listwise methods are considered to be more advanced and accurate as they are specifically designed to solve ranking problems. The objective function is often derived directly from a definition of some ranking measure, e.g. pairwise AUC
or listwise MAP (see \cite{Rendle2009} and \cite{Shi2012} for CP-based and TD-based implementations respectively),
 or constructed in a way that is closely related to those measures \cite{Rendle2010, Rettinger2012}.

These methods typically have a non-trivial loss function with complex data interconections within it which makes it hard to optimize and tune.
In practice, the complexity problem is often resolved with help of handcrafted heuristics and problem-specific constraints (see Sections \ref{subsec:pitf} and \ref{subsec:tfmap}), which simplify the model and improve computational performance.

\section{Tensor-based models in recommender systems}
\label{sec:tensor_models}
Treating data as tensor may bring new levels of flexibility and/or quality into RS models, however there are nuances that should be taken into account and treated properly. This section covers different tensorization techniques used to build advanced RS in various application domains. For all the examples we will use a unified notation (where it is possible) introduced in Section \ref{sec:tensor_intro}, hence it might look different from the notation used in the original papers. This helps to reuse some concepts within different models and build a consistent narrative throughout the text.

\subsection{Personalized search and resource recommendations}
\label{subsec:search}
There is a very tight connection between personalized search and RS. Essentially, recommendations can be considered as a \emph{zero query search} \cite{Allan2012} and, in turn, personalized search engine can be regarded as a query-based RS.

Personalized search systems aim at providing a better search experience by returning the most relevant results, typically web pages (or resources), in response to a user's request. A clicktrough data (i.e. an event log of clicks on the search results after submitting a search query) can be used for this purpose as it contains an information about users' actions and may provide valuable insights into search patterns. The essential part of this data is not just a web page that a user clicks on, but also a context, a query associated with every search request that carries a justification for the user's choice. The utility function in that case can be formulated as:
\begin{displaymath}
    f_R: User \times Resource \times Query \rightarrow Relevance\,Score,
\end{displaymath}
where \emph{Resource} denotes a set of web pages and \emph{Query} is a set of keywords that can be specified by users in order to emphasize their current interests or information needs. In the simplest case a single query can consist of one or a few words (e.g. ``jaguar'' or ``big cat''). More elaborate models could employ additional natural language processing tools in order to breakdown queries into a set of single keywords, e.g. a simple phrase ``what are the colors of the rainbow'' could be transformed into a set \{``rainbow'', ``color''\} and further split into 2 separate queries, associated with the same (\emph{user, resource}) pair.

\subsubsection{CubeSVD}
One of the earliest and at the same time very illustrative works where this formulation was explored with help of tensor factorization is CubeSVD \cite{Sun2005}. The authors build a 3-rd order tensor
$\mathcal{Y} = [y_{ijk}]_{i, j, k=1}^{M, N, K}$.
Values of the tensor represent the level of association (the relevance score) between the user $i$ and the web-page $j$ in presence of the query $k$:
\begin{displaymath}
    \begin{cases}
        y_{ijk} > 0, &\text{ if } (i,j,k) \in S,\\
        y_{ijk} = 0, &\text{ otherwise},
    \end{cases}
\end{displaymath}
where $S$ is an observation history, e.g. a sequence of events described by the triplets (\emph{user, resource, query}). Note that authors in their work use simple queries without processing, e.g. ``big cat'' is a single query term.

The association level can be expressed in various ways, the simplest one is to measure a co-occurrence frequency $f$, e.g. how many times a user has clicked on a specific page after submitting a certain query. In order to prevent an unfair bias towards the pages with high click rates,
it can be restricted to have only values of 0 (no interactions) or 1 (at least one interaction). Or it can be rescaled with a logarithmic function:
\begin{displaymath}
    f' = \log_2(1+f/f_0),
\end{displaymath}
where $f'$ is a new normalized frequency and $f_0$ is, for example, an IDF (Inverse Document Frequency) measure of a web page. Another scaling approach can also be used.

The authors proposed to model the data with a third order TD \eqref{eq:tucker} and in order to find it they applied the HOSVD. Similarly to SVD \eqref{eq:svd},
factors $U \in \mathbb{R}^{M \times r_1}, V \in \mathbb{R}^{N \times r_2}$ and $W \in \mathbb{R}^{K \times r_3}$ represent embedding of users, web pages and queries vectors into a lower-dimensional latent factors space with dimensionalities $r_1, r_2$ and $r_3$ correspondingly.
The core tensor $\mathcal{G} \in \mathbb{R}^{r_1 \times r_2 \times r_3}$ defines the form and the strength of multilinear relations between all three entities in the latent feature space. Once the decomposition is found, the relevance score for any (\emph{user, resource, query}) triplet can be recovered with \eqref{eq:tuckidx}.

With the introduction of new dimensions the data sparsity becomes even higher, which may lead to a numerical instabilities and general failure of the learning algorithm. In order to mitigate that problem, the authors propose several smoothing techniques: based on value imputation with small constant and based on the content similarity of web pages.
They reported an improvement in the overall quality of the model after these modifications.

After applying the decomposition technique the reconstructed tensor $\mathcal{T}$ will contain new non-zero values denoting potential associations between users and web resources influenced by certain queries. The tensor values can be directly used to rank a list of the most relevant resources: the higher the value $t_{ijk}$ is the higher the relevance of the page $j$ to the user $i$ within the query $k$.

This simple TF model does not contain a remedy for some of the typical RS problems such as cold start or real-time recommendations and is most likely to have issues with scalability. Nevertheless, this work is very illustrative and demonstrates the general concepts for building a tensor-based RS.

\subsubsection{TOPHITS}
\label{subsec:tophits}
As has been discussed in Section \ref{subsec:online}, new entities can appear in the system dynamically and rapidly, which in the case of higher order models creates even more computational load, i.e. full recomputation of tensor decomposition quickly becomes infeasible and incremental techniques should be used instead.
However, in some cases simply redefining the model might lower the complexity. As we mentioned in Section \ref{subsec:knn}, a simple approach to reduce the model is to eliminate one of the entities with some sort of aggregation.

For example, instead of considering (\emph{user, resource, query}) triplets we could work with aggregated (\emph{resource, resource, query}) triplets, where every frontal slice $\mathcal{Y}_::k$ of the tensor is simply an adjacency matrix of a resources browsed together under a specific query. Therefore users are no longer explicitly stored and their actions are recorded only in the form of a co-occurrence of resources they searched for.

An example of such a technique is TOPHITS model \cite{Kolda2005, Kolda2006}. This analogy requires an extra explanation as the authors are not modelling users clicking behavior. The model is designed for web-link analysis of a static set of web pages referencing each other via hyperlinks. The data is collected by crawling those web pages and collecting not only links but also keywords associated with them. However, the crawler can be interpreted as a set of users browsing those sites by clicking on the hyperlinked keywords. This draws the connection between CubeSVD and TOPHITS model as the keywords can be interpreted as a short search queries in that case. And, as we stated earlier, users (or crawlers) can be eliminated from the model by constructing an adjacency matrix of linked resources.

The authors of TOPHITS model extend an adjacency matrix of interlinked web pages with the collected keyword information and build a so called adjacency tensor $\mathcal{T} \in \mathbb{R}^{N \times N \times K}$, that encodes hubs, authorities and keywords.
As has been mentioned, the keyword information is conceptually very similar to queries, hence it can be also modelled in a multirelational way.
Instead of TD format the authors prefer to use CP in the form of \eqref{eq:cpbr} with $U, V \in \mathbb{R}^{N \times r}$ and $W \in \mathbb{R}^{K \times r}$ with ALS-based optimization.

The interpretation of this decomposition is different from the CubeSVD. As the authors demonstrate, the weights $\lambda_k$, $(1 \leq k \leq r)$ have a straightforward semantic meaning as they correspond to a set of $r$ specific topics extracted from the overall web page collection.
Accordingly, every triplet of vectors ($u_k, v_k, w_k$) represents a collection of hubs, authorities and keyword terms respectively, characterized by a topic $k$. The elements with higher values in these vectors provide the best-matching candidates under the selected topic,
which allows a better grouping of web pages within every topic and provide means for a personalization.

For example, as the authors show, a personalized ranked list of authorities can be obtained with:
\begin{equation}
    \label{eq:qry}
    \mathbf{a^*} = V \Lambda W^T \mathbf{q},
\end{equation}
where $\Lambda = \diag(\pmb{\lambda})$ is a diagonal matrix and $\mathbf{q}$ is a user-defined query vector of length $K$ with elements $q_t = 1$ if term $t$ belongs to the query and 0 otherwise, $t = \{1, \dots, K$\}. Similarly, a personalized list of hubs can be built simply by substituting factor $V$ with $U$ in \eqref{eq:qry}.

The interpretation of tensor values might seem very natural, however there is an important note to keep in mind. Generally, the restored tensor values might turn both positive and negative. And in most applications the negative values have no meaningful explanation. The non-negative tensor factorization (NTF) \cite{Cichocki2009, Zhou2014} can be employed to resolve that issue (see example in \cite{Chi2010}, and also the connection of NTF to probabilistic factorization model under a specific conditions \cite{Chi2008}).

\subsection{Social tagging}
\label{subsec:tagrec}
A remarkable amount of research is devoted to one specific domain, namely social tagging systems (STS), where predictions and recommendations are based on commonalities in social tagging behavior (also referred as collaborative tagging). A comprehensive overview of the general challenges and state-of-the-art RS methods can be found in \cite{Marinho2010}.

Tags carry a complementary semantic information, that helps STS users to categorize and organize items of their choice. This brings an additional level of interpretation of the user-item interactions, as it exposes the motives behind the user preferences and explains the relevance of particular items.
This observation suggests that tags play an important role in defining the relevance of (\emph{user, item}) pairs, hence all the three entities should be modelled mutually in a multirelational way. The scoring function in that case can be defined as follows:
\begin{displaymath}
    f_R: User \times Item \times Tag \rightarrow Relevance\,Score.
\end{displaymath}
The triplets (\emph{user, item, tag}), coming from an observation history $S$, can be easily translated into a 3rd order tensor $\mathcal{Y} = [y_{ijk}]_{i, j, k=1}^{M, N, K}$, where $M, N, K$ denote the number of users, items and tags respectively. Users are typically not allowed to assign the same tags to the same items more than once, hence the tensor values are strictly binary and defined as:
\begin{displaymath}
    \begin{cases}
        y_{ijk} = 1, &\text{ if } (i,j,k) \in S,\\
        y_{ijk} = 0, &\text{ otherwise.}
    \end{cases}
\end{displaymath}

\subsubsection{Unified framework}
\label{subsec:tag_unified}
As in the case of keywords and queries, tensor dimensionality reduction helps to uncover latent semantic structure of the ternary relations. The values of the reconstructed tensor $\mathcal{T}$ can be interpreted as the likeliness or weight of new links between users, items and tags. These links might be used for building recommendations in various ways: help users assign relevant tags for items \cite{Symeonidis2008a}, find interesting new items \cite{Symeonidis2008b}, or even find like-minded users \cite{Symeonidis2009}. 

The model that is built on top of all three possibilities is described in \cite{Symeonidis2010}. The authors perform a latent semantic analysis on the data with help of the HOSVD. Generally, the base model is similar to CubeSVD (see Section \ref{subsec:search}): items can be treated as resources and tags as queries.

The authors also face the same problem with sparsity. The tensor matricizations $Y_{(n)}, 1 \leq n \leq 3$ within the HOSVD procedure produce highly sparse matrices which may prevent the algorithm from learning the accurate model. In order to overcome that problem they propose a smoothing technique based on a \emph{kernel trick}.


\begin{figure}[bt]
\noindent\centering{
\includegraphics[width=110mm]{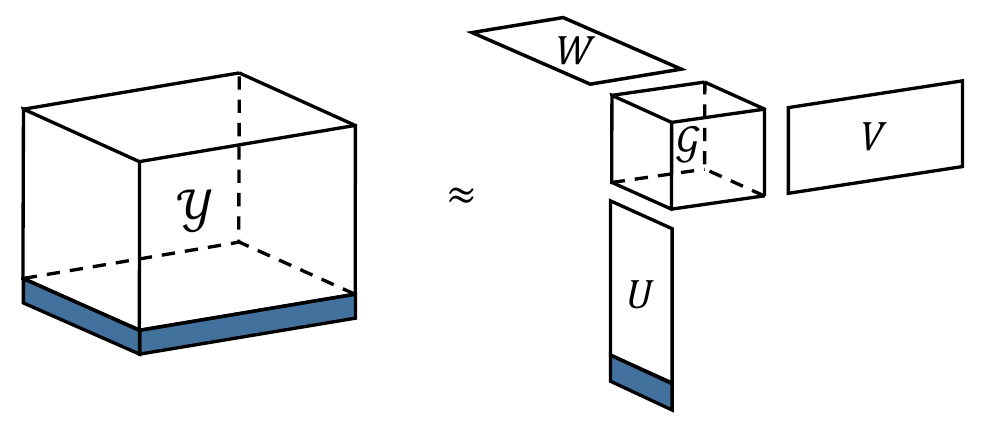}
}
\caption{Higher order folding-in for Tucker decomposition. A slice with new user information in the original data (left) and a corresponding row update of the factor matrix in TD (right) are marked with solid color.}
\label{fig:folding}
\end{figure}

In order to deal with the problem of real-time recommendations (see Section \ref{subsec:online}) the authors adopt a well known \emph{folding-in method} \cite{furnas1988ir} to a higher order case. The folding-in procedure helps to quickly embed a previously unseen entity into the latent features space without recomputing the whole model. For example, an update to a users feature matrix $U$ can be obtained with:
\begin{displaymath}
    \boldsymbol{u}_{new} = \boldsymbol{p} V_1 \Sigma_1^{-1},
\end{displaymath}
where $\boldsymbol{p}$ is a new user information that corresponds to a \emph{row} in the matricitized tensor $Y_{(1)}$; $V_1$ is an already computed (during HOSVD step) matrix of right singular vectors of $T_{(1)}$, and $\Sigma_1$ is a corresponding diagonal matrix of singular values; $\boldsymbol{u}_{new}$ is an \emph{update row} which is appended to the latent factor matrix $U$. The resulting update to reconstructed tensor $\mathcal{T}$ is computed with (see Figure \ref{fig:folding}):
\begin{displaymath}
    \mathcal{T}_{new} = \left[\mathcal{G} \times_2V \times_3W\right] \times_1 \left[
        \begin{array}{c}
            U \\
            \mathbf{u_{new}}
        \end{array}
    \right],
\end{displaymath}
where the term within the left brackets of the right hand side does not contain any new values, e.g. does not require the full recomputation and can be pre-stored, which makes the update procedure much more efficient.

Nevertheless, this typically leads to a loss of orthogonality in factors and negatively impacts the accuracy of the model in the long run. This can be avoided with an \emph{incremental SVD update}, which for the matrices with missing entries was initially proposed by \cite{brand2002incremental}. As the authors demonstrate, it can be also adopted for tensors.

It should be noted, that this is not the only possible option for incremental updates. For example, a different incremental TD-based model with HOOI-based optimization is proposed in \cite{Zhang2014} for a highly dynamic, evolving environment (not related to tag-based recommendations). The authors of this work use an extension of a two-dimensional incremental approach from \cite{ross2008incremental}.

\subsubsection{RTF and PITF}
\label{subsec:pitf}
The models, overviewed so far, has a common ``1/0'' interpretation scheme for a missing values, i.e. all triplets $(i,j,k) \in S$ are assumed to be positive feedback and all others (missing) are negative feedback with zero relevance score. However, as the authors of ranking with TF model (RTF) \cite{Rendle2009} and more elaborate pairwise interaction TF (PITF) \cite{Rendle2010} model emphasize, all missing entries can be split into 2 groups: the true negative feedback and the unknown values. The true negatives
correspond to those triplets of $(i, j, k)$ where the user $i$ has interacted with the item $j$ and has assigned tags different from the tag $k$. More formally, if $P_S$ is a set of all posts that correspond to all observed (\emph{user, item}) interactions, than true negative feedback within any interaction is defined as:
\begin{displaymath}
    Y^{-}_{ij} := \{k | (i, j) \in P_S \land (i, j, k) \notin S\};
\end{displaymath}
trivially, true positive feedback is:
\begin{displaymath}
    Y^{+}_{ij} := \{k | (i, j) \in P_S \land (i, j, k) \in S\}.
\end{displaymath}
All other entries are unknowns and are to be uncovered by the model.

Furthermore, both RTF and PITF models do not require any specific values to be imposed on either known or unknown entries. Instead they only impose pairwise  ranking constraints on the reconstructed tensor values:
\begin{displaymath}
    t_{ijk_1} > t_{ijk_2} \Leftrightarrow (i, j, k_1) \in Y^{+}_{ij} \land (i, j, k_2) \in Y^{-}_{ij}.
\end{displaymath}

These post-based ranking constraints become the essential part of an optimization procedure. The RTF model uses the Tucker format, however
it aims at directly maximizing AUC measure, which, according to the authors, takes the following form:
\begin{displaymath}
    AUC(\theta, i, j) := \frac{1}{|Y^{+}_{ij}| |Y^{-}_{ij}|} \sum_{k^{+} \in Y^{+}_{ij}} \sum_{k^{-} \in Y^{-}_{ij}} \sigma(t_{ijk^{+}} - t_{ijk^{-}}),
\end{displaymath}
where $\theta$ are the parameters of the TD model (as defined in \eqref{eq:opt_general}) and $\sigma(x)$ is a sigmoid function, introduced to make the term differentiable:
\begin{equation}
\label{eq:sigmoid}
    \sigma(x) = \frac{1}{1+e^{-x}}.
\end{equation}

As we are interested in maximizing AUC and due to \eqref{eq:argminj}, the loss function takes the form:
\begin{displaymath}
    L(\mathcal{Y}, \mathcal{T}) = -\sum_{(i,j) \in P_S} AUC(\theta, i, j).
\end{displaymath}
The regularization term of the model is defined by \eqref{eq:opt_reg_cp}.

The authors adopt a stochastic gradient descent (SGD) algorithm for solving the optimization task. However, as they state, directly optimizing the AUC objective is computationally infeasible. Instead, they exploit a smart trick of recombining and reusing precomputed summation terms within the objective and its derivatives. They use this trick for both tasks of learning and building recommendations.

The PITF model is built on top of ideas from RTF model. It adopts Bayesian Personalized Ranking (BPR) technique proposed for MF case in \cite{Rendle2009a} to the ranking approach. The tags rankings for every observed post $(i, j)$ are not deterministically learned like in RTF model but instead are derived from the observations by optimizing the maximum aposteriori estimation. This leads to a similar to RTF optimization objective with similar regularization (excluding the tensor core term which is not present in CP) and slightly different loss function:
\begin{displaymath}
    L(\mathcal{Y}, \mathcal{T}) = - \sum_{(i, j, k_1, k_2) \in D_S} \ln \sigma(t_{ijk_1}-t_{ijk_2}),
\end{displaymath}
where the same notation as in RTF is used; $\sigma$ is a sigmoid function from \eqref{eq:sigmoid} and $D_S$ is a training data, i.e. a set of quadruples:
\begin{equation}
\label{eq:Ds}
    D_S = \{(i, j, k_1, k_2) \,|\, (i, j, k_1) \in S \land (i, j, k_2) \notin S \}.
\end{equation}

An important difference of PITF from RTF is that the complexity of multilinear relations is significantly reduced by leaving only pairwise interactions between all entities. From the mathematical viewpoint it can be considered as a CP model with a special form of partially fixed factor matrices (cf. \eqref{eq:cpidx}):
\begin{equation}
\label{eq:pairwise_full}
    t_{ijk} = \sum_\alpha u_{i\alpha} \, w_{k\alpha}^U + \sum_\alpha v_{j\alpha} \, w_{k\alpha}^V + \sum_\alpha u_{i\alpha} \, v_{j\alpha},
\end{equation}
where $w_{k\alpha}^U$ and $w_{k\alpha}^V$ are the parts of the same matrix $W$ responsible for tags relation to users and items respectively; $u_{i\alpha}$ and $v_{j\alpha}$ are interactional parts of $U$ and $V$.

The authors emphasize, that the user-item interaction term does not contribute to the BPR-based ranking optimization scheme which yields even more simple equation, that becomes an essential part of the PITF model:
\begin{equation}
\label{eq:pairwise}
    t_{ijk} = \sum_\alpha u_{i\alpha} \, w_{k\alpha}^U + \sum_\alpha v_{j\alpha} \, w_{k\alpha}^V.
\end{equation}

Another computational trick that helps to train the model even faster without sacrificing the quality is random sampling within the SGD routine. All the quadruples in $D_S$ corresponding to a post $(i, j)$ are highly overlapped with respect to the tags associated with them. Therefore, learning with some randomly sampled quadruples is likely to have a positive effect on learning the remaining ones.

In order to verify the correctness and effectiveness of such simplifications the authors conduct experiments with both BPR-tuned TD and CP and demonstrate that PITF algorithm achieves close or even better quality of recommendations while learning features faster than the other two TF methods.

Despite its computational effectiveness, the original PITF model is lacking the support for the real-time recommendation scenarios, where rebuilding the full model for each new user, item or tag could be prohibitive. The authors of \cite{Liao2014} overcome this limitation by introducing the folding-in procedure compatible with the PITF model and demonstrate its ability to provide high recommendations quality. Worth noting here, that a number of variations of the folding-in technique are available for different TF methods, see \cite{Zhang2011a}. 

The idea of modelling higher order relations in a joint pairwise manner similar to \eqref{eq:pairwise} has been explored in various application domains and is implemented in various settings, either straightforwardly or as a part of a more elaborate RS model \cite{Gantner2010, Hidasi2014, zhao2015crafting, Shan2016}.
There are several generalized models \cite{Wermser2011, Rettinger2012}, \cite{Hidasi2015}, that also use this idea. They are covered in more details in Sections \ref{subsec:CARTD} and \ref{subsec:GFF} of this work.


\subsubsection{Improving the prediction quality}
As has been already mentioned in Section \ref{subsec:tag_unified} high data sparsity typically leads to a less accurate predictive models. This problem is common across various RS domains. Another problem, specific to STS, is tag ambiguity and redundancy. The following are the examples of some of the most common techniques, developed to deal with these problems.

The authors of CubeRec \cite{Xu2006} propose a clustering-based separation mechanism. This mechanism builds clusters of triplets (\emph{user, item, tag}) based on the proximity of tags derived from the item-tag matrix. With this clustering some of the items and tags can belong to several clusters at the same time, according to their meaning. After that the initial problem is split into a number of sub-problems (corresponding to clusters) of a smaller size and hence, with a more dense data. Every subproblem is then factorized with the HOSVD similarly to \cite{Sun2005}, and the resulting model is constructed as a combination of all the smaller TF models.

The authors of the clustering-based TD model (ClustHOSVD) \cite{Symeonidis2015} also employ clustering approach. However, instead of splitting the problem, they replace tags by tag clusters and apply the HOOI method (which is named AlsHOSVD by the authors) directly to the modified data consisting of (\emph{user, item, tag cluster}) triplets. They also demonstrate the effect of different clustering techniques on the quality of RS.

As can be seen, many models benefit from clustering either prior to or after the factorization step. This suggests that it can also be beneficial to perform simultaneous clustering and factorization. This idea is explored by the authors of \cite{fu2015joint}, where they demonstrate the effectiveness of such an approach.


A further improvement can be achieved with hybrid models (see Section \ref{subsec:model_based}), that exploit a content information and incorporate it into a tensor-based CF model. It should be noted, however, that there is no ``single-bullet'' approach, suitable for all kinds of problems, as it highly depends on the type of data used as a source of content information.

The authors of \cite{Nanopoulos2010} exploit acoustic features for music recommendations in a tag-based environment. The features, extracted with specific audio-processing techniques, are used to measure the similarity between different music samples. The authors make an assumption that similarly sounding music is likely to have similar tags, which allows to propagate tags to the music that was not tagged yet. With this assumption the data is augmented with new triplets of (\emph{user, item, tag}), which leads to a more dense data and results in a better predictive quality of the HOSVD model.

The TF and tag clustering (TFC) model \cite{Rafailidis2013} combines both content exploitation and tag clustering techniques. The authors focus on the image recommendations problem, thus they use an image processing techniques in order to find items' similarities and propagate highly relevant tags. Once the tag propagation is completed, the authors find tag clusters (naming them topics) and build new association triplets  (\emph{user, item, topic}), which are further factorized with the HOSVD.

As a last remark in this section, the idea of model splitting, proposed in  the CubeRec model, was also explored in a more general setup in \cite{Wang2013}. The authors consider a multiple context environment, where user-item interactions may depend on various contexts such as location, time, activity, etc. This is generally modelled with an $N$-th order tensor, where $N > 3$. Instead of dealing with higher number of dimensions and greater sparsity, the authors propose to build a separate model for every context type, which transforms the initial problem into a collection of a smaller problems of order 3. Then all the resulting TF models are combined with specific weights (based on the context influence measure proposed by the authors) and can be used to produce recommendations. However, despite the ability to better handle the sparsity issue, the model may loose some valuable information about the relations between different types of context. A more general methods for multi-context problems are covered in Section \ref{subsec:CARS}.

%
%
%

\subsection{Temporal models}
\label{subsec:temporal}
User consumption patterns may change in time. For example, the interest of TV users may correlate not only with a topic of a TV program, but also with a specific time of the day. In retail user preferences may vary depending on the season.
Temporal models are designed to learn those time-evolving patterns in data by taking the time aspect into account, which can be formalized with the following way scoring function:
\begin{displaymath}
    f_R: User \times Item \times Time \rightarrow Relevance\,Score.
\end{displaymath}

Even though the general problem statement looks already familiar, when working with the $Time$ domain one should mind the difference between the evolving and periodic (e.g. seasonal) events which may require a special treatment.

\subsubsection{BPTF}
\label{subsec:BPTF}
One of the models that exploits periodicity of events is the Bayesian Probabilistic TF (BPTF) \cite{Xiong2010}. It uses seasonality to reveal trends in retail data and predict the orders that will arrive in the ongoing season based on the season's start and previous purchasing history. The key feature of the model is the ability to produce forecasts on the sales of the new products, that were not present in previous seasons. The model captures dynamic changes in both product designs and customers' preferences solely from the history of transactions and does not require any kind of an expert knowledge.

The authors develop a probabilistic latent factors model by introducing priors on the parameters; i.e. the latent feature vectors are allowed to vary and the variance of relevance scores is assumed to follow a Gaussian distribution:
\begin{displaymath}
    t_{ijk} | U, V, W \sim \mathcal{N}(<U_{i:},V_{j:},W_{k:}>, \gamma^{-1}),
\end{displaymath}
where $\gamma$ is an observations precision and $<U_{i:},V_{j:},W_{k:}>$ denotes a right hand side of \eqref{eq:cpidx}. Note, that in the original work the authors use a transposed version of the factor matrices, e.g. any column of the factor $U$ in their work represents a single user, the same holds for two other factors.

In order to prevent the overfitting the authors also impose prior distributions on $U$ and $V$:
\begin{gather*}
    U_{i:} \sim \mathcal{N}(0, \sigma_U^2 I),\\
    V_{j:} \sim \mathcal{N}(0, \sigma_V^2 I).\\
\end{gather*}
Furthermore, the formulation for the time factor $W$ takes into account its evolving nature and implies smooth changes in time:
\begin{gather*}
    W_{k:} \sim \mathcal{N}(W_{k-1:}, \sigma_{dW}^2 I),\\
    W_{0:} \sim \mathcal{N}(\mu_W, \sigma_{0}^2 I).
\end{gather*}
The time factor $W$ rescales the user-item relevance score with respect to the time-evolving trends and the probabilistic formulation helps to account for the users who do not follow those trends.

The authors show that maximizing the log-posterior distribution $\log p(U,V,W, W_{0,:}|\mathcal{Y})$ with respect to $U,V,W$ and $W_{0:}$ is equivalent to an  optimization task with the weighted square loss function:
\begin{equation}
\label{eq:square_loss}
    L(\mathcal{Y}, \mathcal{T}) = \sum_{(i, j, k) \in S} (y_{ijk} - t_{ijk})^2,
\end{equation}
and a bit more complex regularization term:
\begin{displaymath}
    \Omega(\theta) = \frac{\lambda_U}{2} \|U\|_F^2 + \frac{\lambda_V}{2} \|V\|_F^2 + \frac{\lambda_{dT}}{2}\sum_{k=1}^K \|W_{k:}-W_{k-1:}\|^2 + \frac{\lambda_{0}}{2} \|W_{0:} - \mu_W\|^2,
\end{displaymath}
where $\lambda_{U} = (\alpha \sigma_{U})^{-1}, \lambda_{V} = (\alpha \sigma_{V})^{-1}, \lambda_{dW} = (\alpha \sigma_{dW})^{-1}, \lambda_{0} = (\alpha \sigma_{0})^{-1}$ and the last two terms are due to a dynamic problem formulation.
The number of parameters of this model makes the task of optimization almost infeasible. However, the authors come up with an elaborate MCMC-based integration approach, that makes the model almost parameter-free and also scales well.

\subsubsection{TCC}
The authors of  TF-based subspace clustering and preferences consolidation model (TCC) \cite{Wang2016} exploit the periodicity in usage patterns of the IPTV users in order to, at first, identify them and, secondly, provide with more relevant recommendations, even if those users share the same IPTV account (for example, across all family members). This gives a slightly different definition of a utility function:
\begin{displaymath}
    f_R: Account \times Item \times Time \rightarrow Relevance\,Score,
\end{displaymath}
where $Account$ is the domain of all registered accounts and the number of accounts is not greater than the number of users, i.e. $|Account| \leq |User|$.
Initial tensor $\mathcal{Y}$ is built from the triplets (\emph{account, item, time}) and its values are just the play counts.

In order to be able to find a correct mapping of the real users to the known accounts, the authors introduce a concept of a \emph{virtual user}.
Within the model the real user is assumed to be a composition of particular virtual users $u_{ak}$ which express the specific user's preferences tied to a certain time periods, e.g.:
\begin{displaymath}
    u_{ak} := \{(a, p_k) \,|\, a \in A, p_k \in P, p_k \neq \varnothing \},
\end{displaymath}
where $a$ is an account from the set of all accounts $A$, $p_k$ is a sub-period from the set of all non-overlapping time periods $P$.

As the authors state, manually splitting the time data into the time slots $p_k$ does not fit the data well and they propose to find those sub-periods from the model. They first solve the SGD-based optimization task \eqref{eq:opt_general}, \eqref{eq:argminj} for the TD with the same weighted squared loss function as in \eqref{eq:square_loss} and regularization term as in \eqref{eq:opt_reg} (with $\lambda_G = \lambda_U = \lambda_V = \lambda_W = \frac{1}{2}\lambda$).
Once the model factors are found, the sub-periods $p_k$ can be obtained by clustering the time feature vectors:
\begin{displaymath}
    P \leftarrow \text{ k-Means clustering of the rows of } W.
\end{displaymath}

Then the consolidation of virtual users into the real ones can be done in 2 steps. At first, a binary similarity measure is computed between different pairs of virtual users ($u_{ak}$, $u_{ak'}$) corresponding to the same account $a$. The second step is to combine similar virtual users so that every real user is represented as a set of virtual ones. This is done with help of a graph-based techniques.
Once the real users are identified, recommendations can be produced with a user-based kNN approach.
As the authors demonstrate, the proposed method not only provides a tool for user identification, but also outperforms standard kNN and TF-based methods applied without any prior clustering.

\subsection{General context-aware models}
\label{subsec:CARS}
In previous sections we have discussed TF methods targeted at specific classes of problems: keyword- or tag-based recommendations, temporal models. They all have one thing in common - the use of a third entity leading to a higher level of granularity and better predictive capabilities of a model. This leads to an idea of generalization of such an approach, that is suitable for any model formulated in the form of \eqref{eq:NDutility}.
\subsubsection{Multiverse}
One of the first attempts towards this generalization is the Multiverse model \cite{Karatzoglou2010}. The authors define context as any set of variables that influence users' preferences and propose to model it by the $N$-th order TD with $N\!-\!2$ contextual dimensions:
\begin{displaymath}
    \mathcal{T} = [\![\mathcal{G}; U, V, W_1, W_2, \dots, W_{N-2}]\!],
\end{displaymath}
where factors $W_i, i\!=\!\{1, \dots,  N\!-\!2\}$ represent a corresponding embedding of every contextual variable into a reduced latent space and all factors including $U$ and $V$ are not restricted to be orthogonal.
As the authors state, the model is suitable for any contextual variables over a finite categorical domain. It should be noted, that the main focus of the work is systems with an explicit feedback and the model is optimized for the error-based metrics, which does not guarantee an optimal items ranking as has been discussed in Section \ref{subsec:eval}.

Following the general form of an optimization objective stated in \eqref{eq:opt_general}, the authors use the weighted loss function:
\begin{displaymath}
    L(\mathcal{T}, \mathcal{Y}) = \frac{1}{\|\mathcal{G}\|_1}\sum_{(i,j,k) \in S} l(t_{ijk}, \ y_{ijk}),
\end{displaymath}
where $l(t_{ijk}, \ y_{ijk})$ is a pointwise loss function, that can be based on $l_2$, $l_1$ or other types of distance measure. The example is provided for the 3rd order case, however, it can be easily generalized to a higher orders.
The authors also use the same form of the regularization term as in \eqref{eq:opt_reg}, as it enables trivial optimization procedure.

In order to fight against the growing complexity for the higher order cases they propose a modification of the SGD algorithm. Within a single optimization step the updates are performed on every row of the latent factors independently. For example, an update for $i$-th row of $U$:
\begin{displaymath}
    U_{i:} \leftarrow U_{i:} - \eta \, \lambda_U U_{i:} - \eta \, \partial_{t_{ijk}} l(t_{ijk}, \ y_{ijk})
\end{displaymath}
is independent on all other factors and thus all the updates can be performed in parallel. The parameter $\eta$ defines the model's learning step.

In addition to the general results on the real dataset, this work features a comprehensive experimentation on the semi-synthetic data that shows the impact of a contextual information on the RS models performance.
It demonstrates that high context influence leads to better quality of the selected context-aware methods, among which the proposed TF approach gives the best results, while a context-unaware method's quality significantly degrades.


\subsubsection{TFMAP}
\label{subsec:tfmap}
Similarly to the previously discussed PITF model, the TF for MAP optimization model (TFMAP) \cite{Shi2012} also targets optimal ranking, however it exploits the MAP metric instead of AUC.
The model is designed for an implicit feedback systems which means that the original tensor $\mathcal{Y}$ is binary with non-zero elements reflecting the fact that the interaction has occurred:
\begin{equation}
\label{eq:binary}
    y_{ijk} =
    \begin{cases}
        1, &\text{if } (i, j, k) \in S, \\
        0, &\text{otherwise}.
    \end{cases}
\end{equation}

The optimization objective is drawn from the MAP definition:
\begin{displaymath}
    MAP = \frac{1}{MK} \sum_{i=1}^M \sum_{k=1}^K
     \frac{\sum_{j=1}^N \frac{y_{ijk}}{r_{ijk}} \sum_{j'=1}^N y_{ij'k} \, \mathbb{I}\left(r_{ij'k} \leq r_{ijk}\right)}
    {\sum_{j=1}^N y_{ijk}},
\end{displaymath}
where $r_{ijk}$ denotes the rank of the item $j$ in the items list of the user $i$ under the context type $k$ and $\mathbb{I}\left(\cdot\right)$ is an indicator function, which is equal to 1 if the condition is satisfied and 0 otherwise, both depend on the reconstructed values of $\mathcal{T}$. In order to make the metric smooth and differentiable the authors propose two approximations:
\begin{align*}
    \frac{1}{r_{ijk}} &\approx \sigma(t_{ijk}), \\
    \mathbb{I}\left(r_{ij'k} \leq r_{ijk}\right) &\approx \sigma(t_{ij'k} - t_{ijk}),\\
\end{align*}
where $t_{ijk}$ is calculated with \eqref{eq:cpidx} (which makes the model a CP-based) and $\sigma$ is a sigmoid function defined by \eqref{eq:sigmoid}. Notably, $t_{ij'k} - t_{ijk} = <U_{i:},V_{j':} - V_{j:},W_{k:}>$, where we use the same notation as in BPTF model, see Section \ref{subsec:BPTF}.

The model also follows the standard optimization formulation stated in \eqref{eq:opt_general}, where the loss function is just a negative MAP gain, i.e. $L(\mathcal{T}, \mathcal{Y}) = - MAP$, and the regularization has the form of \eqref{eq:opt_reg_cp}.

Note, that MAP optimization also has a weighted form due to \eqref{eq:binary}, however, the computation complexity would still be prohibitively high due to its complex structure.
In order to mitigate that, the authors propose the fast learning algorithm: for each (\emph{user, context}) pair only a limited set of a representative items (a buffer) is considered, which in turn, allows to control the computational complexity. They also provide an efficient algorithm of sampling the ``right'' items and constructing the buffer, which does not harm the overall quality of the model.

%
%
%
%
%
%

\subsubsection{CARTD}
\label{subsec:CARTD}
The CARTD model (Context-Aware Recommendation Tensor Decomposition) \cite{Wermser2011, Rettinger2012}
provides a generalized framework for an arbitrary number of contexts and also targets an optimal ranking instead of a rating prediction. Under the hood the model extends the BPR-based ranking approach used in the PITF model to the higher order cases.

The authors introduce a unified notion of an entity.
A formal task is to find the list of the most relevant entities within a given contextual situation.
Remarkably, all the information, that is used to make recommendations more accurate and relevant, is defined as a context. In that sense, not only information like tag, time, location, user attributes, etc. is considered to be a context, even users themselves might be defined as a context of an item. This gives a more universal formulation for the recommendations task:
\begin{equation}
    f_R: Entity \times Context_1 \times \ldots \times Context_n \rightarrow Relevance\,Score.
\end{equation}

As an illustration to that, a quadruple (\emph{user, item, time, location}) maps to (\emph{context$_1$, entity, context$_2$, context$_3$}). Obviously, the definition of the entity depends on the task. For example, in case of social interactions prediction with (\emph{user, user, attribute}) triplets, the main entity as well as one of the context variables will be a user.

The observation data in a typical case of a user-item interactions can be encoded similarly to \eqref{eq:Ds}:
\begin{displaymath}
    D_S := \{ (e, f, c_1, \dots, c_n) \ | \ (e, c_1, \dots, c_n) \in S \land (f, c_1, \dots, c_n) \notin S\},
\end{displaymath}
where $e$ and $f$ are the entities (i.e. items) and $c_i, i = \{1, \dots, n\}$ denotes a context type (includes users). As the authors emphasize, this leads to a huge sparsity problem, and instead they propose to relax conditions and instead build the following set for learning the model:
\begin{displaymath}
    D_A := \{ (e, f, c_1, \dots, c_n) \ | \ \forall c_i: \#_{c_i}(e) > \#_{c_i}(f)\},
\end{displaymath}
where $\#_{c_i}(\cdot)$ indicates the number of occurrences of an entity within the context $c_i$. The rule $\#_{c_i}(e) > \#_{c_i}(f)$ denotes the prevalence of the entity $e$ over the entity $f$ with respect to all possible contexts.

The optimization objective will also look similar to the one used in the PITF model with the loss function defined as:
\begin{displaymath}
    L(\mathcal{T}, \mathcal{Y}) = - \sum_{(e, f, c_1, \dots, c_n) \in D_A} \text{ln } \sigma \left(t_{\{c\},e}-t_{\{c\},f}\right),
\end{displaymath}
where $\{c\}$ denotes a set of all context variables $c_i$ and the tensor values $t_{ijk}$ are calculated with help of the reduced CP model with the pairwise only interactions, similarly to \eqref{eq:pairwise}:
\begin{displaymath}
    t_{\{c\}, e} = \sum_{i=1}^n {v}_e^{E, C_i} {v}_{c_i}^{C_i, E},
\end{displaymath}
where ${v}_{e}^{E, C_i}$ and ${v}_{c_i}^{C_i, E}$ are the elements at the cross section of the $e$-th row and the $i$-th column of the factor matrices $V^{E, C_i}$ and $V^{C_i, E}$ respectively.
As in the previous cases, the regularization term $\Omega(\theta)$ have similar to \eqref{eq:opt_reg_cp} form, which includes all the factors from $\theta$:
\begin{displaymath}
    \theta = \{V^{E, C_1}, V^{C_1, E}, \dots, V^{E, C_n}, V^{C_n, E}\}.
\end{displaymath}


\subsubsection{iTALS}
As has been mentioned in the introduction (see Section \ref{subsec:intro_implicit}), an implicit feedback does not always correlate with the actual user preferences, thus a simple binary scheme (as in \eqref{eq:binary}) may not be accurate enough. For this reason, the authors of the iTALS model (ALS-based implicit feedback recommender algorithm) \cite{Hidasi2012} propose to use the confidence-based interpretation of an implicit feedback introduced in \cite{Hu2008} and adopt it for the higher order case.

They introduce the dense tensor $\mathcal{W}$ that assigns non-zero weights for both observed and unobserved interactions. For the $n$-th order tensor it has the following form:
\begin{equation}
\label{eq:weights}
    \begin{cases}
        w_{i_1, \ldots, i_n} = \alpha \cdot \#(i_1, \ldots, i_n), &\text{ if } (i_1, \ldots, i_n) \in S,\\
        w_{i_1, \ldots, i_n} = 1, &\text{ otherwise},
    \end{cases}
\end{equation}
where $\#(i_1, \ldots, i_n)$ is the number of occurrences of the tuple $(i_1, \ldots, i_n)$ (e.g. the combination of the user $i_1$ and the item $i_2$ interacted within the set of contexts $i_3, \ldots, i_n$) in the observation history; $\alpha$ is set empirically and $\alpha \cdot \#(i_1, \ldots, i_n) > 1$ which means that the observed events provide more confidence in the user preferences than the unobserved ones.

The loss function will then take the form:
\begin{displaymath}
    L(\mathcal{T}, \mathcal{Y}) = \sum_{i_1, \ldots, i_n} w_{i_1, \ldots, i_n} (t_{i_1, \ldots, i_n} - y_{i_1, \ldots, i_n})^2,
\end{displaymath}
where weights $w_{i_1, \ldots, i_n}$ are defined by \eqref{eq:weights}, $y_{i_1, \ldots, i_n}$ are the values of a binary feedback tensor of order $n$, defined  similarly to \eqref{eq:binary}, and $t_{i_1, \ldots, i_n}$ are the values of the reconstructed tensor.

The model uses CP with an ALS-based optimization procedure and a standard regularization similar to \eqref{eq:opt_reg_cp}. The latent feature vectors are encoded in the rows of the factor matrices, not the columns, i.e. following the authors' notation, we should rewrite \eqref{eq:cp1} as:
\begin{displaymath}
    \mathcal{T} = [\![M_1^T, \dots, M_n^T ]\!],
\end{displaymath}
where $M_i \ (1 \leq i \leq n)$ are transposed factors of the CP decomposition.

The authors show, how an efficient computation over the dense tensor can be achieved with the same tricks, that are used in \cite{Hu2008} for the matrix case. The model also has a number of modifications \cite{Hidasi2013}: based on the conjugate gradient approach (iTALS-CG) and the coordinates descent approach (iTALS-CD) where an additional features compression is achieved by the Cholesky decomposition. This makes the iTALS-CD model to learn even faster than MF methods. While performing on approximately the same level of accuracy as the state-of-the-art Factorization Machines (FM) method \cite{rendle2011fast}, it is capable of learning more complex latent relations structure. Another modification is the pairwise ``PITF-like'' reduction model, named iTALSx \cite{Hidasi2014}.

\subsubsection{GFF}
\label{subsec:GFF}
The General Factorization Framework (GFF) \cite{Hidasi2015} further develops the main ideas of the family of iTALS models. Within the GFF model different CP-based factorization models (also called a preference models) are combined in order to capture the intrinsic relations between users and items influenced by an arbitrary number of contexts. As in many other works the authors of GFF model fix the broad definition of the context as an entity, which ``value is not determined solely by the user or the item'', i.e. not a content information (see Section \ref{subsec:incontext}).

The model can be better explained with the example. Let us consider the problem of learning the scoring function as follows:
\begin{equation}
    f_R: U \times I \times S \times Q \rightarrow Relevance\,Score,
\end{equation}
where $U$ and $I$ are the domains of \emph{users} and \emph{items} respectively; $S$ stands for \emph{season} and denotes the periodicity of the events (see Section \ref{subsec:temporal}); $Q$ describes the sequential consumption patterns, e.g. what are the previous items that were also consumed with the current one (see \cite{Hidasi2012} for broader set of examples). Let us also define the pairwise interactions between users and items as $UI$ (standard CF model), between items and seasons as $IS$ and so forth. Using the same notation we can also define multi-relational interactions, such as $UIS$ for a 3-way user-item-season interactions or $UISQ$ for the 4-way interactions between all 4 types of entities.

In total, there could be 2047 different combinations of interactions, yet not all of them are feasible in terms RS model, as not all of them may contribute to the preference model.

As the result, GFF generates a very flexible multirelational model that allows to pick the most appropriate scheme of interactions, that does not explode the complexity of the model and meanwhile achieves a high quality of recommendations. Based on the experiments the authors conclude: ``leaving out useless interactions results in more accurate models''.



\begin{table}
    \centering{
    \resizebox{\textwidth}{!}{
    \begin{tabular}[h]{l|lllllcl}
        Name & Type & Algorithm & Domain & Entities & Optimization  &
        \parbox{2cm}{\centering Ranking \\ prediction}    & Online \\
    \hline
        TOPHITS \cite{Kolda2005},       2005 &     CP & ALS   & Link prediction     & Resources, Keyword    &  pointwise &   Yes &    No \\
        CubeSVD \cite{Sun2005},         2005 &     TD & HOSVD & Personalized Search & User, Resource, Query &  pointwise &   Yes &    No \\
            RTF \cite{Rendle2009},      2009 &     TD & SGD   & Folksonomy          & User, Item, Tag       &  pairwise  &   Yes &    No \\
           BPTF \cite{Xiong2010},       2010 &     CP & MCMC  & Temporal            & User, Item, Time      &  pointwise &    No &    No \\
     Multiverse \cite{Karatzoglou2010}, 2010 &     TD & SGD   & Context-awareness   & User, Item, Contexts  &  pointwise &    No &    No \\
           PITF \cite{Rendle2010},      2010 & CP$^*$ & SGD   & Folksonomy          & User, Item, Tag       &  pairwise  &   Yes &    No \\
   TagTR \cite{Symeonidis2010},         2010 &     TD & HOSVD & Folksonomy          & User, Item, Tag       &  pointwise &   Yes &   Yes \\
          TFMAP \cite{Shi2012},         2012 &     CP & SGD   & Context-awareness   & User, Item, Context   &  listwise  &   Yes &    No \\
          CARTD \cite{Rettinger2012},   2012 & CP$^*$ & SGD   & Context-awareness   & Item, Contexts        &  pairwise  &   Yes &    No \\
     ClustHOSVD \cite{Symeonidis2015},  2015 &     TD & HOOI  & Folksonomy          & User, Item, Tag       &  pointwise &   Yes &    No \\
            GFF \cite{Hidasi2015},      2015 & CP$^*$ & ALS   & Context-awareness   & User, Item, Contexts  &  pointwise &   Yes &    No \\

    \end{tabular}
    }
    }

\caption{Comparison of popular TF methods.
The \emph{Ranking prediction} column shows whether a method is evaluated against ranking metrics.
The \emph{Online} column denotes a support for real-time recommendations for new users (e.g. folding-in).
\newline
\newline
$^{*}$ Method uses pairwise reduction concept, initially introduced in PITF.}

\label{tab:methods}
\end{table}

We have reviewed so far a diverse set of tensor-based recommendation techniques. Clearly, tensors help represent and model complex environments in a convenient and congruent way, suitable for various problem formulations. Nevertheless, as we have already stated earlier, the most common practical task for RS is to build a ranked list of recommendations (a top-$n$ recommendations task). In this regard, we summarize related features of some of the most illustrative in our opinion methods in Table \ref{tab:methods}. We also take into account a support for real-time scenarios in dynamic environments.

\subsection{Other models}
\label{subsec:other}
Unfortunately, it is almost impossible to review all available TF models from various domains. The flexibility that comes with the tensor-based formulation provides means for limitless combinations of various RS settings and models.
Here we briefly describe some of them, that were not referenced yet, but have an interesting application and/or implementation.

\paragraph{Social interactions.}
The authors of \cite{Kutty2012} focus on recommending new connections for users in specific communities like online dating or professional networks
They emphasize that typically there are two types of groups of people (for instance, employee and employer) in job seeking networks. In order to account for that and avoid unnecessary recommendations within the same group (e.g. employer-employer) they split the problem into two parallel subproblems corresponding to each individual group and model it with the CP. The final result is than aggregated from both subproblems by selecting only those predicted links (i.e. recommendations) which are present in both groups.


The TOPHITS approach, described in Section \ref{subsec:tophits}, is shown to be applicable for the authorities ranking task in the Twitter community \cite{Sizov2010}. This technique  can be potentially used for improving the followee recommendations for twitter users.

\paragraph{Social tagging.}
A few works for image tagging \cite{Panagopoulos2015, Barmpoutis2015} use a interesting representation of data, initially proposed in \cite{dunlavy2011multilinear}. Users and images, uploaded by them in social network, are encoded together into a single long vector. These vectors are used to build a set of adjacency matrices, that are made with respect to certain conditions and then stacked together in a tensor.
With this approach, every frontal slice of the tensor describes different kinds of relations: friendship relations between users, user-image connections, tag relations for both users and items, etc.

\paragraph{Temporal models.}

The authors of \cite{Yao2015} add a so called \emph{social regularization}, introduced in \cite{ma2011recommender}, into a standard optimization routine. The idea behind this modification is to use not only a ``wisdom of crowds'' like in standard CF approach, but also utilize information about social relationships (i.e. friendship) of the user in order to bring more trust into the recommendations an improve the overall accuracy.

The work \cite{Misaghian2013} combines both social tagging and temporal models. The authors build a 4-th order tensor from (\emph{user, item, tag, time}) quadruples and decompose it with the HOSVD. In order estimate relevance scores and recommend new items for users, they first summarize values, corresponding to the observations, over the third (tag) mode.

An interesting hybrid approach for modelling user preferences dynamics is proposed in \cite{Rafailidis2014a}. The authors build a tensor from (\emph{user, item, time-period}) triplets and combine it with an auxiliary content information (user attributes) with help of a \emph{coupled tensor-matrix factorization method} \cite{Acar2011, Ermi2013}. The idea of coupled tensor-matrix factorizations provides an additional level of flexibility in model construction and is used in various RS domains with complex setup \cite{Yang2014, Bhargava2015}.

%

\paragraph{Multi-criteria ratings systems.}
The authors of \cite{Wang2012} explore the rich sentiment information in a product reviews. They extract opinions from text and craft a multi-aspect (or multi-criteria, see \cite{Adomavicius2011a}) ratings system on top of it. This data is used to build a third order tensor in the form (\emph{user, item, aspect}), with tensor values denoting the ratings within each aspect (including the explicit ratings). A CP-based factorization model is used to reconstruct missing values and predict unknown ratings more accurately.

A similar idea of multi-criteria ratings model was also used as a part of a sophisticated model in \cite{Nilashi2014}. However, the authors did not have to do any text analysis as the aspect data was populated by users themselves and provided within the dataset. They also applied the HOSVD method instead of the CP.

\paragraph{Mobility and geolocation.}
Modern social networks allow to share not only a content, such as images or videos, but also link that content to a specific locations using the Global Positioning System (GPS) services. With the broad access of mobile devices to the internet, this provides rich information about user interests and behavior and allows building highly personalized context-aware services and applications. For example, the authors of \cite{Symeonidis2011, Symeonidis2013} model location-based user activities with a third order tensor (\emph{user, activity, location}) for providing locations and activities recommendations. The authors of \cite{Sattari2014} use the tensor for the personalized activities rating prediction. These works use the Tucker tensor format and apply the HOSVD for its reconstruction.

\paragraph{Cross-domain recommendations.}
Another interesting direction is combining a cross-domain knowledge, e.g. user consumption patterns of books, movies, music, etc., in order to improve recommendations quality. Knowledge about the patterns from one domain may help to build more accurate predictions in another (this is a so called \emph{knowledge transfer} learning). Moreover, modelling these cross-domain relations mutually may also help to achieve a higher recommendations quality across all domains. An interesting challenge in these tasks is a varying number of items in different domains, which requires a special treatment. A few notable and quite different techniques of the tensor-based knowledge transfer learning are proposed in \cite{Chen2013} and \cite{HuL.CaoJ.XuG.CaoL.GuZ.&Zhu2013}.





\paragraph{Special factorization methods.}
In the theory of matrix approximations there is well known pseudo-skeleton decomposition method \cite{Goreinov1997}, that allows to use only a small sample of the original matrix elements in order to find an approximate matrix factorization within the desired accuracy. This result is shown to be generalizable to a higher order case \cite{Oseledets2008, Oseledets2010}, and, remarkably, is especially suitable for sparse data. The main benefit of such a sampling approach is the reduced factorization complexity in terms of both the number of operations and the number of elements required for computation, which is especially advantageous in case of tensor-based models. A special case of such a class of TF algorithms is used in the TensorCUR model \cite{Mahoney2008} for product recommendations.

\section{Conclusion}
\label{sec:conclusion}
In this survey we have attempted to overview a broad range of tensor-based methods used in recommender systems to date. As we have seen, these methods provide powerful set of tools for merging various types of additional information, that increases flexibility, customizability and quality of recommendation models. Tensorization enables creative and non-trivial setups, going far beyond standard user-item paradigm, and finds its applications in various domains. Tensor-based models can also be used as a part of more elaborate systems, providing compressed latent representations as an input for other well-developed techniques.

One of the main concerns for the higher order models is inevitable growth of computational complexity with increasing number of dimensions. Even for mid-sized production systems, that have to deal with highly dynamic environments, this might have negative implications, such as inability to produce recommendations for new users instantly,  in a timely manner. This type of issues can be firmly addressed with incremental update and higher order folding-in techniques.
The former allow to update the entire model, while performing computations only on new data. The latter allows to calculate recommendations in cases when new data is already present in the system but was not yet included into the model.

Despite the encouraging results, there is an issue related to the applicability of CP and TD decompositions. When the number of dimensions becomes much higher than 3, application of TD-based methods becomes infeasible due to explosion of storage requirements. On the other side, CP is generally ill-posed which may potentially lead to numerical instabilities. A possible cure for this problem is to use TT/HT decomposition. In our opinion, this is a promising direction for further investigations.
\section{Acknowledgements}
The authors would like to thank Maxim Rakhuba and Alexander Fonarev for their help for improving the manuscript, and also Michael Thess for insightful conversations.
\bibliographystyle{abbrv}
\bibliography{literature}
\end{document}